\documentclass[11pt,letterpaper]{article}

\usepackage[utf8]{inputenc}
\usepackage[T1]{fontenc}
\usepackage{amsmath,amssymb,amsthm}
\usepackage{algorithm}
\usepackage{algpseudocode}
\usepackage{graphicx}
\usepackage[numbers]{natbib}
\usepackage{booktabs}
\usepackage{listings}
\usepackage{xcolor}
\usepackage{enumitem}
\usepackage{tikz}
\usetikzlibrary{shapes.geometric,arrows.meta,positioning,fit,backgrounds,calc,decorations.pathreplacing,patterns}
\usepackage{pgfplots}
\pgfplotsset{compat=1.18}
\usepackage{geometry}
\usepackage[bookmarks=true]{hyperref}

\hypersetup{
  colorlinks=true,
  linkcolor=blue!60!black,
  citecolor=green!50!black,
  urlcolor=blue!70!black,
  pdfauthor={Mohsen Arjmandi},
  pdftitle={Sensi: Learn One Thing at a Time---Curriculum-Based Test-Time Learning for LLM Game Agents},
  pdfkeywords={LLM agents, test-time learning, curriculum learning, ARC-AGI, game playing}
}

\title{Sensi: Learn One Thing at a Time---Curriculum-Based Test-Time Learning for LLM Game Agents}

\author{
  Mohsen Arjmandi\thanks{Correspondence: \texttt{mohsen.arjmandi@gmail.com}. Blog: \url{https://freddiespirit.substack.com} (Freddie Spirit).}\\
  Independent Researcher (CTO, evolutionID)
}

\date{}

\begin{document}

\maketitle

\begin{abstract}
Large language model (LLM) agents deployed in unknown environments must learn task structure at test time, but current approaches require thousands of interactions to form useful hypotheses. We present \textbf{Sensi}, an LLM agent architecture for the ARC-AGI-3 game-playing challenge that introduces structured test-time learning through three mechanisms: (1)~a two-player architecture separating perception from action, (2)~a curriculum-based learning system managed by an external state machine, and (3)~a database-as-control-plane that makes the agent's context window programmatically steerable. We further introduce an LLM-as-judge component with dynamically generated evaluation rubrics to determine when the agent has learned enough about one topic to advance to the next. We report results across two iterations: Sensi~v1 solves 2 game levels using the two-player architecture alone, while Sensi~v2 adds curriculum learning and solves 0 levels---but completes its entire learning curriculum in approximately 32 action attempts, achieving 50--94$\times$ greater sample efficiency than comparable systems that require 1,600--3,000 attempts. We precisely diagnose the failure mode as a self-consistent hallucination cascade originating in the perception layer, demonstrating that the architectural bottleneck has shifted from \emph{learning efficiency} to \emph{perceptual grounding}---a more tractable problem.
\end{abstract}

\section{Introduction}
\label{sec:intro}

The ARC-AGI-3 challenge~\citep{arcprize2024arcagi} presents a distinctive test of machine intelligence: agents must play pixel-art puzzle games with no prior knowledge of the rules, mechanics, or win conditions. Unlike standard game-playing benchmarks where the rules are specified or can be inferred from a large training corpus, ARC-AGI-3 requires agents to discover the game's dynamics entirely through interaction---a form of test-time learning that mirrors how humans approach unfamiliar situations.

Current approaches to this challenge are remarkably sample-inefficient. Systems such as Agentica from Symbolica reportedly require 1,600 to 3,000 game interactions to build a working model of a single game's mechanics. This inefficiency fundamentally limits the practical applicability of LLM agents in settings where test-time compute budgets are constrained, a concern that has become increasingly relevant as the field explores test-time scaling~\citep{wei2022chain,wang2023selfconsistency,yuksekgonul2025tttdiscover}.

We introduce \textbf{Sensi}, an LLM agent that evolves across two design iterations to address this sample efficiency problem. The core architectural insight is separation of concerns: perception should be decoupled from action, and learning should be structured as a curriculum rather than an undirected search.

\textbf{Sensi~v1} splits the agent into two communicating roles---an Observer that maintains hypotheses about the game world, and an Actor that selects actions to test those hypotheses. This two-player architecture, where cognition is factored into perception and decision-making, enables structured hypothesis accumulation. Using ChatGPT~5.1 as the backbone, v1 solves 2 levels of the ARC-AGI-3 challenge with the notable property that pass@10 equals pass@1: results are consistently reproducible given the same accumulated knowledge.

\textbf{Sensi~v2} adds three mechanisms on top of the two-player foundation: (i)~a \emph{curriculum} that orders learning goals into a sequential queue with a state machine managing progression, (ii)~a \emph{database-as-control-plane} where the agent's entire cognitive state resides in SQLite tables that are programmatically injected into prompts each turn, and (iii)~a \emph{sense scorer} implementing LLM-as-judge~\citep{zheng2024judging} with dynamically generated evaluation rubrics. V2 solves 0 levels---an honest negative result---but completes its full learning curriculum in approximately 32 interactions, a 50--94$\times$ improvement in sample efficiency.

The failure mode is itself a contribution: we identify a \emph{self-consistent hallucination cascade} where errors in the perception layer (frame differencing) propagate through the hypothesis pipeline, producing internally coherent but factually incorrect game models that receive high scores from the sense evaluator. This precisely localizes the bottleneck: the architecture works; the perception does not.

Our contributions are:
\begin{enumerate}[leftmargin=2em]
  \item A \textbf{two-player architecture} that separates perception from action via distinct Observer and Actor LLM roles communicating through structured hypothesis lists (\S\ref{sec:v1}).
  \item A \textbf{curriculum learning system} with an external state machine that sequences learning goals and accumulates verified knowledge as facts for subsequent learning items (\S\ref{sec:v2}).
  \item A \textbf{database-as-control-plane} pattern that makes the LLM's context window externally programmable through database state, enabling modular and steerable agent behavior (\S\ref{sec:database}).
  \item An \textbf{LLM-as-judge with dynamic rubrics} for evaluating learning progress, where both the evaluation metric and the scoring are performed by separate LLM calls (\S\ref{sec:sense}).
  \item A \textbf{sample efficiency analysis} demonstrating 50--94$\times$ improvement over baselines, together with a precise diagnosis of the hallucination cascade failure mode (\S\ref{sec:results}).
\end{enumerate}

\section{Preliminaries}
\label{sec:prelim}

\subsection{ARC-AGI-3 Environment}
\label{sec:arcagi}

ARC-AGI-3~\citep{arcprize2024arcagi} is a game-based evaluation where agents play pixel-art puzzle games without prior knowledge of the rules. At each timestep $t$, the agent observes a game frame represented as a 3D tensor:
\begin{equation}
  \mathcal{F}_t \in \mathbb{Z}_{16}^{L \times H \times W}
\end{equation}
where $L$ is the number of grid layers, $H$ and $W$ are the spatial dimensions (up to $64 \times 64$), and each cell takes an integer value in $[0, 15]$ representing a color. The agent selects from a discrete action space:
\begin{equation}
  \mathcal{A} = \{a_0, a_1, \ldots, a_7\}
\end{equation}
where $a_0$ is \textsc{reset} (restarts the game), $a_1$ through $a_5$ are simple actions (directional movement, interact), $a_6$ is a coordinate-parameterized action taking $(x, y)$ arguments, and $a_7$ is an additional simple action. After executing action $a_t \in \mathcal{A}$, the agent receives the next frame $\mathcal{F}_{t+1}$ and a scalar score $s_t \in [0, 254]$, along with a status code from $\{\texttt{NOT\_PLAYED}, \texttt{NOT\_FINISHED}, \texttt{GAME\_OVER}, \texttt{WIN}\}$.

Crucially, the semantics of each action are \emph{not} provided to the agent. The agent must discover what each action does, how the game mechanics work, and what constitutes winning---all from raw interaction.

\subsection{POMDP Formulation}
\label{sec:pomdp}

The ARC-AGI-3 problem can be formulated as a partially observable Markov decision process (POMDP)~\citep{kaelbling1998planning} with an important distinction: the agent must discover not only the state dynamics but also the reward function itself. Let $\langle \mathcal{S}, \mathcal{A}, T, R, \Omega, O \rangle$ denote the POMDP, where $\mathcal{S}$ is the (hidden) game state space, $\mathcal{A}$ is the action space, $T: \mathcal{S} \times \mathcal{A} \to \Delta(\mathcal{S})$ is the unknown transition function, $R: \mathcal{S} \times \mathcal{A} \to \mathbb{R}$ is the unknown reward function, $\Omega$ is the observation space (rendered frames), and $O: \mathcal{S} \to \Delta(\Omega)$ is the observation function.

The agent's task is two-fold: (1)~learn an approximate model of $T$ and $R$ from interactions, and (2)~use this model to select actions that maximize the game score. This dual-discovery requirement distinguishes ARC-AGI-3 from standard POMDP settings where the dynamics are assumed known or learnable from a fixed distribution.

\subsection{Test-Time Learning}
\label{sec:ttl}

It is important to situate Sensi within the landscape of test-time adaptation methods. We distinguish three paradigms:

\textbf{Test-time compute (TTC).} Methods such as chain-of-thought
prompting~\citep{wei2022chain} and
self-consistency~\citep{wang2023selfconsistency} allocate additional
inference-time computation to improve output quality. The model's
parameters remain frozen; only the reasoning trajectory is extended.

\textbf{Test-time training (TTT).} Methods such as TTT~\citep{sun2024learning} and TTT-Discover~\citep{yuksekgonul2025tttdiscover} perform gradient-based updates to the model's parameters on each test instance. The model literally learns---in the weight-update sense---at test time.

\textbf{Test-time in-context learning.} Sensi occupies a third position: the model's parameters remain frozen, but the agent accumulates structured knowledge in an external database that is injected into the context window each turn. This is neither pure inference (the agent's effective knowledge base changes) nor gradient-based training (no parameters are updated), extending in-context test-time learning ideas from frameworks like EvoTest~\citep{he2025evotest}. We call this \emph{in-context knowledge accumulation with persistent structured state}. The key distinction from standard in-context learning is that the accumulated knowledge persists across context windows through the database, enabling learning trajectories that span hundreds of turns without being limited by context length.

\section{Sensi v1: Two-Player Architecture}
\label{sec:v1}

The foundational insight of Sensi is that a single LLM call asked to simultaneously perceive a game state, reason about what changed, maintain hypotheses, and select an action performs poorly at all four tasks. V1 addresses this by splitting cognition into two cooperating roles.

\subsection{Observer and Actor}

Each turn in Sensi v1 involves two LLM calls:
\begin{align}
  \text{Player}_1(\mathcal{F}_t, \mathcal{F}_{t-1}, a_{t-1}, \Delta_t) &\;\rightarrow\; (G_t, K_t) \label{eq:player1} \\
  \text{Player}_2(G_t, K_t) &\;\rightarrow\; (d_t, a_t) \label{eq:player2}
\end{align}
where $G_t$ is the \emph{guesses} list (hypothesized game mechanics), $K_t$ is the \emph{figured-out} list (confirmed observations), $d_t \in \{\textsc{guess}, \textsc{informed}\}$ is the decision type indicating whether the action is exploratory or knowledge-based, and $\Delta_t$ is a visual description of changes between consecutive frames.

The prompt frames this as cooperative play: ``you and your friend are playing a game together where you see the board and your friend chooses an action.'' This framing leverages the LLM's capacity for perspective-taking and collaborative reasoning.

\subsection{Stochastic Exploration via Confidence-Modulated Agency}
\label{sec:stochastic}

A distinctive aspect of v1's design is its treatment of \emph{confidence as a dynamic variable} that modulates the agent's exploration behavior. Rather than using fixed exploration schedules (as in $\epsilon$-greedy strategies) or curiosity-driven bonuses~\citep{pathak2017curiosity,schmidhuber1991curious}, Sensi instructs the LLM to act based on what has been figured out---allowing the model to dynamically modulate its behavior based on its evolving epistemic state.

This design exploits an underappreciated capacity of LLMs: the ability to model \emph{stochastic aspects of human agency}. When humans face genuinely unknown situations, their behavior is not purely logical---it includes intuitive leaps, confidence-weighted guesses, and exploratory randomness that collectively help navigate unfamiliar territory. By representing confidence dynamically in the agent's state (through the distinction between guesses and figured-out items), Sensi allows the LLM to approximate this stochastic exploration naturally, without requiring explicit exploration mechanisms.

Formally, the agent's update rule can be expressed as:
\begin{align}
  \text{Sense}(t+1) &= \text{Sense}(t) + \text{LLM}(\text{Sense}(t), S(t+1)) \\
  S(t+1) &\sim P(\cdot \mid S(t), A(t)) \\
  A(t+1) &= \text{LLM}(\text{Sense}(t+1))
\end{align}
where $\text{Sense}(t) = (G_t, K_t)$ is the evolving epistemic state. Unlike thinking models that generate stochastic exploration through random token sampling, Sensi's exploration emerges from the \emph{structure of the epistemic state}: when the guesses list is large relative to the figured-out list, the Actor naturally behaves more exploratorily; as items migrate from guesses to figured-out, behavior becomes more directed.

\subsection{Results and Limitations}

Sensi v1 was evaluated on ARC-AGI-3 using ChatGPT~5.1 as the backbone model. It successfully solved the first 2 levels of game LS20, discovering 15 correct facts about the game mechanics including player movement rules, energy consumption, key-door interactions, and the win condition.

A notable property of v1's results is that \textbf{pass@10 = pass@1}: given the same figured-out list, the Actor consistently selects the correct action sequence. This contrasts with thinking-model approaches where results vary across runs due to sampling stochasticity, and suggests that the two-player architecture produces more deterministic and reproducible behavior.

However, v1 has fundamental limitations: (i)~there is no mechanism to control \emph{what} the agent learns or in what order, (ii)~there is no way to verify whether the agent has actually learned something correctly, and (iii)~at level~3, the agent entered a degenerate state where it stopped updating its understanding and repeated single actions across 100+ turns. These limitations motivate the structured learning approach of v2.

\section{Sensi v2: Curriculum Learning with Programmable Context}
\label{sec:v2}

The core idea behind v2 is to impose structure on the learning process itself: \textbf{learn one thing at a time}. Instead of allowing the agent to explore an unstructured hypothesis space, v2 introduces a queue of learning items that the agent must master sequentially, with an external state machine managing progression and an LLM-based judge verifying learning outcomes.

\begin{figure}[t]
  \centering
  \resizebox{\textwidth}{!}{%
  \begin{tikzpicture}[
    llmbox/.style={rectangle, draw=blue!70!black, fill=blue!8, thick, minimum width=2.1cm, minimum height=1.1cm, align=center, rounded corners=3pt, font=\small},
    dbbox/.style={cylinder, draw=orange!70!black, fill=orange!8, thick, shape border rotate=90, aspect=0.25, minimum width=2.4cm, minimum height=1.6cm, align=center, font=\scriptsize},
    iobox/.style={rectangle, draw=gray!60, fill=gray!8, thick, minimum width=1.6cm, minimum height=0.8cm, align=center, rounded corners=2pt, font=\small},
    smbox/.style={rectangle, draw=green!60!black, fill=green!8, thick, minimum width=1.5cm, minimum height=0.7cm, align=center, rounded corners=2pt, font=\scriptsize},
    arr/.style={-{Stealth[length=5pt]}, thick, color=blue!50!black},
    darr/.style={-{Stealth[length=5pt]}, thick, color=orange!70!black, dashed},
    label/.style={font=\scriptsize\itshape, text=gray!70!black},
  ]
    \node[iobox] (frame) at (0,0) {$\mathcal{F}_t, \mathcal{F}_{t-1}$};

    \node[llmbox] (fd) at (2.8,0) {\textbf{FrameDiff}\\[-2pt]{\scriptsize LLM$_1$}};
    \node[llmbox] (mg) at (5.6,0) {\textbf{MetricGen}\\[-2pt]{\scriptsize LLM$_2$}};
    \node[llmbox] (ss) at (8.4,0) {\textbf{SenseScore}\\[-2pt]{\scriptsize LLM$_3$}};
    \node[llmbox] (p1) at (11.2,0) {\textbf{Player$_1$}\\[-2pt]{\scriptsize LLM$_4$}};
    \node[llmbox] (p2) at (14.0,0) {\textbf{Player$_2$}\\[-2pt]{\scriptsize LLM$_5$}};

    \node[iobox] (action) at (16.4,0) {$a_t$};

    \draw[arr] (frame) -- (fd);
    \draw[arr] (fd) -- node[above, label] {$\Delta_t$} (mg);
    \draw[arr] (mg) -- node[above, label] {$\mu_{I^*}$} (ss);
    \draw[arr] (ss) -- node[above, label] {$\phi_t, r_t$} (p1);
    \draw[arr] (p1) -- node[above, label] {$G_t, K_t$} (p2);
    \draw[arr] (p2) -- (action);

    \node[dbbox] (db) at (8.4,-2.8) {\textbf{SQLite DB}\\[1pt]\texttt{\scriptsize items\_to\_learn}\\[-1pt]\texttt{\scriptsize game\,/\,inputs}\\[-1pt]\texttt{\scriptsize guesses}\\[-1pt]\texttt{\scriptsize figured\_outs}};

    \draw[darr] (db.north) -- (ss.south);
    \draw[darr] (db) -| node[below left, label, pos=0.25] {\scriptsize facts, history} (p1.south);
    \draw[darr] (db) -| node[below left, label, pos=0.15] {\scriptsize facts} (p2.south);
    \draw[darr] (p1.south) ++(0.15,0) -- ++(0.15,-0.8) -- ++(0.6,0) node[right, label] {\scriptsize write $G_t,K_t$} |- (db.east);

    \node[label, text=red!60!black] at (5.6,0.85) {once per item};

    \node[smbox] (nr) at (3.0,-2.8) {\texttt{not\_reached}};
    \node[smbox, fill=yellow!15] (lr) at (3.0,-4.0) {\texttt{learning}};
    \node[smbox, fill=green!15] (co) at (3.0,-5.2) {\texttt{completed}};

    \draw[arr, green!60!black] (nr) -- node[right, font=\scriptsize] {pick} (lr);
    \draw[arr, green!60!black] (lr) -- node[right, font=\scriptsize] {$\phi_t \geq \tau$} (co);
    \draw[arr, green!60!black] (lr.west) -- ++(-0.5,0) |- node[left, font=\scriptsize, pos=0.25] {$\phi_t < \tau$} (lr.west);

    \draw[-{Stealth[length=5pt]}, thick, green!60!black, dashed] (co.east) -- ++(1.5,0) node[right, font=\scriptsize, text=green!50!black] {$K_t \to$ facts $\to$ DB};

    \node[font=\footnotesize\bfseries, text=blue!60!black] at (8.4,1.2) {LLM Pipeline (per turn)};
    \node[font=\footnotesize\bfseries, text=green!50!black] at (3.0,-1.9) {State Machine};

  \end{tikzpicture}%
  }
  \caption{\textbf{Sensi v2 architecture.} Each turn involves up to five LLM calls orchestrated through a pipeline. The SQLite database serves as the control plane: all agent state resides in database tables that are queried to construct prompts and updated with each turn's outputs. The curriculum state machine (left) manages learning progression, promoting figured-out items to facts when a learning item is completed. MetricGen (LLM$_2$) runs only once when a new learning item is activated.}
  \label{fig:v2_arch}
\end{figure}
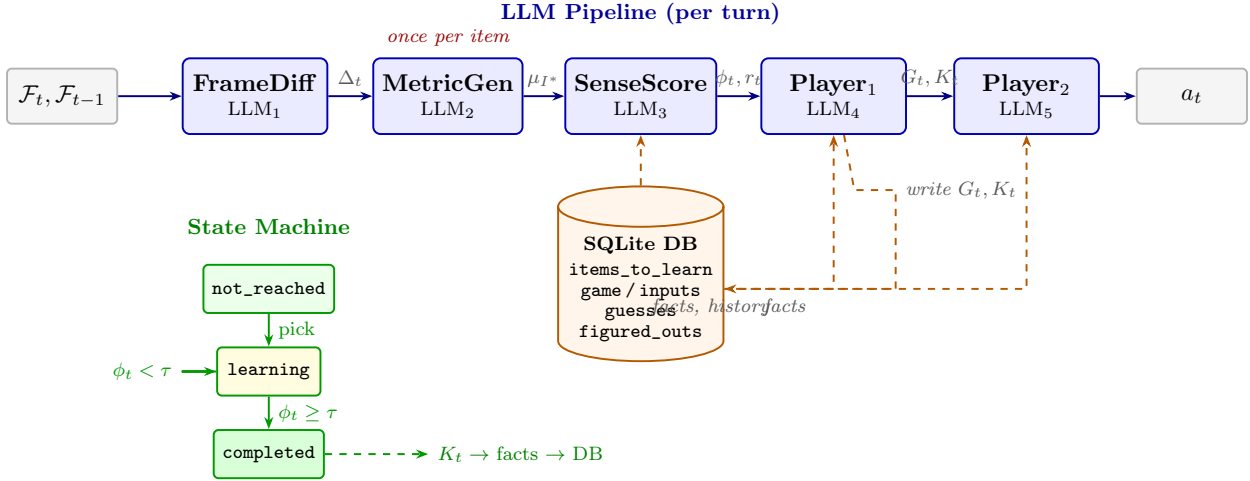

\subsection{Turn Architecture}
\label{sec:turn}

Each turn in Sensi v2 involves up to five LLM calls organized as a pipeline:
\begin{equation}
  \underbrace{\text{FrameDiff}}_{\text{LLM}_1} \;\rightarrow\; \underbrace{\text{MetricGen}}_{\text{LLM}_2} \;\rightarrow\; \underbrace{\text{SenseScore}}_{\text{LLM}_3} \;\rightarrow\; \underbrace{\text{Player}_1}_{\text{LLM}_4} \;\rightarrow\; \underbrace{\text{Player}_2}_{\text{LLM}_5}
  \label{eq:pipeline}
\end{equation}

\textbf{Frame Differencing ($\text{LLM}_1$).} The raw grid $\mathcal{F}_t$ is rendered to a scaled pixel image via a color map ($\mathbb{Z}_{16} \to \text{RGB}$). Two consecutive frames are passed to a multimodal LLM that returns a structured JSON diff:
\begin{lstlisting}[caption={Frame differencing module.},label=lst:framediff]
def frame_diff_finder(self, current_frame, prev_frame) -> str:
    prediction = self.frame_diff_module(
        prev_frame=DSPyImage(url=encode_image(prev_frame)),
        current_frame=DSPyImage(url=encode_image(current_frame)),
    )
    return prediction.diff_json  # structured JSON
\end{lstlisting}
The diff schema captures added, removed, and moved objects, UI changes, and a high-level summary.
This gives the Observer a structured description of \emph{what changed} rather than forcing it to compare raw pixels.

\textbf{Metric Generation ($\text{LLM}_2$).} Called once per learning item when it enters the \texttt{learning} state. Generates an evaluation rubric for the sense scorer (detailed in \S\ref{sec:sense}).

\textbf{Sense Scoring ($\text{LLM}_3$).} Called every turn for the active learning item. Produces a score $\phi_t \in [1, 10]$ and reasoning text (detailed in \S\ref{sec:sense}).

\textbf{Player 1---Observer ($\text{LLM}_4$).} Receives the game frame, frame diff, previous guesses and figured-out lists, accumulated facts, the current learning target, sense score feedback, and action history. Outputs updated guesses and figured-out lists. The prompt includes guidelines on list curation: keep valid items, edit with new information, remove contradicted items, and promote confirmed guesses to figured-out.

\textbf{Player 2---Actor ($\text{LLM}_5$).} Receives guesses, figured-out lists, facts, and the current learning target. Outputs a decision type ($d_t \in \{\textsc{guess}, \textsc{informed}\}$) and exactly one action. The Actor's objective is to resolve the Observer's uncertainties by choosing actions that help discriminate between competing hypotheses.

\subsection{Learning Queue and State Machine}
\label{sec:curriculum}

At the start of each game, Sensi v2 initializes a learning curriculum stored in an SQLite table:

\begin{lstlisting}[caption={Curriculum initialization.},label=lst:curriculum]
def initialize_items_to_learn(self, game_id, card_id):
    # Facts: already-known truths
    fact_items = [
        "RESET starts the game",
        "all available actions: ACTION1, ..., ACTION7, RESET",
    ]
    # Learning items: to be mastered sequentially
    default_items = [
        "learn what each action does in the game",
        "learn how actions affects your energy while playing",
        "learn how to win the game",
    ]
    for item_name in default_items:
        cur.execute("""
            INSERT OR IGNORE INTO items_to_learn
            (game_id, card_id, item_name, state, threshold)
            VALUES (?, ?, ?, 'not_reached', 8)
        """, (game_id, card_id, item_name))
\end{lstlisting}

Each learning item follows a three-state lifecycle managed by a state machine:

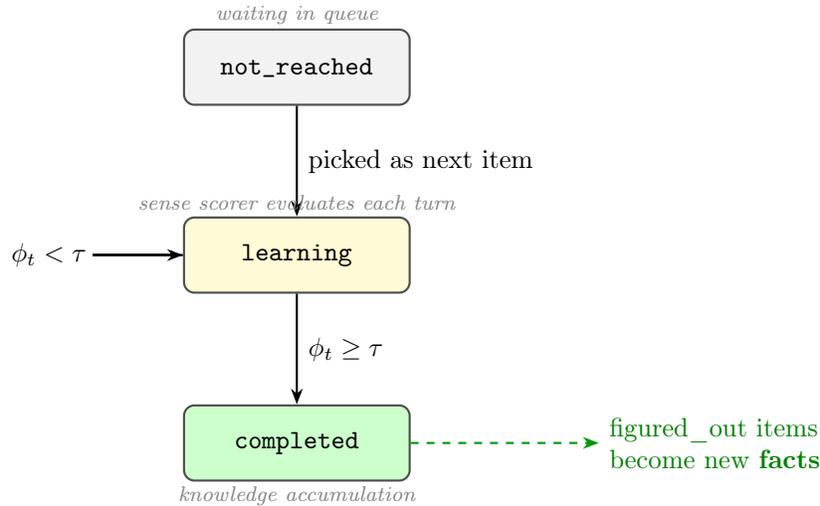
\begin{figure}[t]
  \centering
  \begin{tikzpicture}[
    state/.style={rectangle, draw=black!70, thick, minimum width=3cm, minimum height=1cm, align=center, rounded corners=4pt, font=\small\ttfamily},
    arr/.style={-{Stealth[length=6pt]}, thick},
    lbl/.style={font=\small, fill=white, inner sep=2pt},
  ]
    \node[state, fill=gray!10] (nr) at (0,0) {not\_reached};
    \node[state, fill=yellow!20] (lr) at (0,-2.5) {learning};
    \node[state, fill=green!20] (co) at (0,-5.0) {completed};

    \draw[arr] (nr) -- node[lbl, right, xshift=2pt] {picked as next item} (lr);
    \draw[arr] (lr) -- node[lbl, right, xshift=2pt] {$\phi_t \geq \tau$} (co);

    \draw[arr] (lr.west) -- ++(-1.2,0) |- node[lbl, left, pos=0.25] {$\phi_t < \tau$} (lr.west);

    \draw[arr, dashed, green!60!black] (co.east) -- ++(2.5,0) node[right, font=\small, align=left, text=green!50!black] {figured\_out items\\become new \textbf{facts}};

    \node[font=\scriptsize\itshape, text=gray] at (0,0.7) {waiting in queue};
    \node[font=\scriptsize\itshape, text=gray] at (0,-1.8) {sense scorer evaluates each turn};
    \node[font=\scriptsize\itshape, text=gray] at (0,-5.7) {knowledge accumulation};

  \end{tikzpicture}
  \caption{\textbf{Learning item state machine.} Each curriculum item progresses through three states. The self-loop on \texttt{learning} represents repeated evaluation until the sense score meets the threshold $\tau$. Upon completion, accumulated figured-out items are promoted to facts, creating a knowledge accumulation chain.}
  \label{fig:state_machine}
\end{figure}

Formally, let $\mathcal{I} = \{I_1, I_2, \ldots, I_n\}$ be the ordered sequence of learning items, each with state $\sigma_i \in \{\texttt{not\_reached}, \texttt{learning}, \texttt{completed}, \texttt{fact}\}$. At any turn $t$, the \emph{active item} is:
\begin{equation}
  I^* = \min_{i} \{I_i \mid \sigma_i \in \{\texttt{learning}, \texttt{not\_reached}\}\}
  \label{eq:active}
\end{equation}
When $I^*$ transitions to \texttt{completed}, its accumulated figured-out entries are promoted to facts:
\begin{equation}
  \sigma_{I^*} = \texttt{completed} \;\Longrightarrow\; \forall\, k \in K_t : \sigma_k \leftarrow \texttt{fact}
  \label{eq:promotion}
\end{equation}
These new facts become part of the context for subsequent learning items, creating a \textbf{knowledge accumulation chain}: each learning item builds upon the verified (or putatively verified) knowledge from all previous items.

\subsection{Sense Scoring: LLM-as-Judge for Learning Progress}
\label{sec:sense}

How does the agent know when it has learned enough about a topic? V2 introduces a \emph{sense scorer}---a separate LLM call acting as an external judge. Critically, the judge does not use a fixed rubric. Instead, the rubric itself is dynamically generated by another LLM call, creating a two-phase evaluation system.

\textbf{Phase 1: Metric Generation.} When a new learning item becomes active, a \texttt{Metric\-Generator\-Signature} is invoked to produce a verification criterion:
\begin{lstlisting}[caption={Dynamic metric generation via DSPy.},label=lst:metricgen]
class MetricGeneratorSignature(dspy.Signature):
    """Given an item the agent needs to learn about a game,
    generate a metric to verify that the item has been learned."""

    item_to_learn: str = dspy.InputField(
        desc="The item/concept the agent needs to learn"
    )
    learning_metric: str = dspy.OutputField(
        desc="A criteria and description that a judge will use
         to give a score on how good a grasp the learner has
         on the item. Score between 1 to 10."
    )
\end{lstlisting}

\clearpage
\textbf{Phase 2: Sense Scoring.} On each subsequent turn, the \texttt{SenseScorerSignature} evaluates the agent's current understanding against the generated metric:
\begin{lstlisting}[caption={Sense scoring via DSPy.},label=lst:sense]
class SenseScorerSignature(dspy.Signature):
    """Score the agent's understanding of a learning item."""

    item_to_learn: str = dspy.InputField()
    learning_metric: str = dspy.InputField()
    facts: List[str] = dspy.InputField()
    figured_out: List[str] = dspy.InputField()

    sense_score: int = dspy.OutputField(
        desc="Score from 1-10 indicating learning progress."
    )
    reasoning: str = dspy.OutputField(
        desc="Brief explanation for the score"
    )
\end{lstlisting}

Let $\mu_i$ denote the metric generated for item $I_i$, and $\phi_t$ the sense score at turn $t$. The transition condition is:
\begin{equation}
  \phi_t(I^*, \mu_{I^*}, \mathcal{K}_t, K_t) \;\geq\; \tau_{I^*} \;\;\Longrightarrow\;\; \sigma_{I^*} \leftarrow \texttt{completed}
  \label{eq:transition}
\end{equation}
where $\mathcal{K}_t$ is the accumulated fact set and $\tau_{I^*}$ is the threshold (default $\tau = 8$ out of 10). The sense reasoning $r_t$ is fed back to Player~1 on the next turn, providing explicit feedback about \emph{why} the score was what it was and what knowledge is still missing.

\subsection{Database as Control Plane}
\label{sec:database}

One of the most architecturally significant choices in v2 is using an SQLite database as the agent's \emph{programmable context}. Every turn, the agent reads its state from and writes its state to six database tables (Table~\ref{tab:schema}).

\begin{table}[t]
  \centering
  \caption{\textbf{Database schema.} Six SQLite tables constitute the agent's externalized cognitive state.}
  \label{tab:schema}
  \begin{tabular}{ll}
    \toprule
    \textbf{Table} & \textbf{Purpose} \\
    \midrule
    \texttt{items\_to\_learn} & Learning curriculum with state machine \\
    \texttt{inputs} & Key-value store for per-turn game state \\
    \texttt{game} & Turn-by-turn history of frames, actions, diffs \\
    \texttt{guesses} & Player 1's hypothesis lists per turn \\
    \texttt{figured\_outs} & Player 1's confirmed observations per turn \\
    \texttt{losing\_action\_seqs} & Action sequences that led to game over \\
    \bottomrule
  \end{tabular}
\end{table}

This is not mere persistence---it is a \textbf{control plane}. By modifying database contents, one changes the agent's behavior without altering code or prompts. Want the agent to skip action learning? Delete those rows. Want to seed it with human-provided facts? Insert them. Want to change the learning order? Reorder the items.

The LLM's context window becomes programmable through database state:
\begin{equation}
  \text{Prompt}_t = f\!\left(\mathcal{F}_t,\; \text{DB}.\texttt{facts},\; \text{DB}.\texttt{item\_to\_learn},\; \text{DB}.\texttt{history},\; \text{DB}.\texttt{sense\_score}\right)
  \label{eq:prompt}
\end{equation}

This architecture differs fundamentally from standard prompt engineering or template-based agent frameworks. In those approaches, the prompt structure is static and the content varies only through direct substitution. In Sensi v2, the database mediates between the symbolic control structure (curriculum, state machine) and the neural inference engine (LLM), creating a form of \emph{neuro-symbolic programming} where the symbolic layer guides the neural layer through structured learning.

The practical implication is \emph{external steerability}: the entire learning process can be monitored, modified, and debugged through standard database operations. This is analogous to how control planes in distributed systems~separate the data path (what the system does) from the control path (how the system is configured).

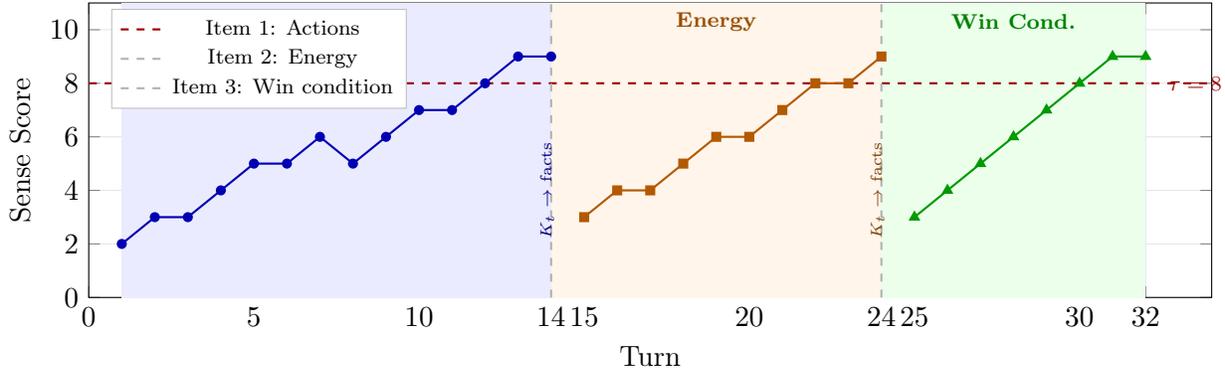
\begin{figure}[t]
  \centering
  \begin{tikzpicture}
    \begin{axis}[
      width=\textwidth,
      height=5.5cm,
      xlabel={Turn},
      ylabel={Sense Score},
      xmin=0, xmax=34,
      ymin=0, ymax=11,
      ytick={0,2,4,6,8,10},
      xtick={0,5,10,14,15,20,24,25,30,32},
      grid=major,
      grid style={gray!20},
      legend style={at={(0.02,0.98)}, anchor=north west, font=\scriptsize, draw=gray!50},
      clip=false,
    ]

    \fill[blue!8] (axis cs:1,0) rectangle (axis cs:14,11);
    \fill[orange!8] (axis cs:14,0) rectangle (axis cs:24,11);
    \fill[green!8] (axis cs:24,0) rectangle (axis cs:32,11);

    \addplot[dashed, thick, red!60!black, domain=0:34] {8};
    \node[font=\scriptsize, text=red!60!black] at (axis cs:33.5,8) {$\tau=8$};

    \addplot[dashed, thick, gray!60] coordinates {(14,0) (14,11)};
    \addplot[dashed, thick, gray!60] coordinates {(24,0) (24,11)};

    \addplot[thick, blue!70!black, mark=*, mark size=1.5pt] coordinates {
      (1,2) (2,3) (3,3) (4,4) (5,5) (6,5) (7,6) (8,5) (9,6) (10,7) (11,7) (12,8) (13,9) (14,9)
    };

    \addplot[thick, orange!70!black, mark=square*, mark size=1.5pt] coordinates {
      (15,3) (16,4) (17,4) (18,5) (19,6) (20,6) (21,7) (22,8) (23,8) (24,9)
    };

    \addplot[thick, green!60!black, mark=triangle*, mark size=1.8pt] coordinates {
      (25,3) (26,4) (27,5) (28,6) (29,7) (30,8) (31,9) (32,9)
    };

    \node[font=\scriptsize\bfseries, text=blue!60!black] at (axis cs:7.5,10.3) {Actions};
    \node[font=\scriptsize\bfseries, text=orange!60!black] at (axis cs:19,10.3) {Energy};
    \node[font=\scriptsize\bfseries, text=green!50!black] at (axis cs:28,10.3) {Win Cond.};

    \node[font=\tiny, text=blue!50!black, rotate=90, anchor=south] at (axis cs:14.4,4) {$K_t \to$ facts};
    \node[font=\tiny, text=orange!50!black, rotate=90, anchor=south] at (axis cs:24.4,4) {$K_t \to$ facts};

    \legend{Item 1: Actions, Item 2: Energy, Item 3: Win condition}

    \end{axis}
  \end{tikzpicture}
  \caption{\textbf{Sense score progression over turns} (illustrative). The agent's sense score for each curriculum item rises toward the threshold $\tau = 8$ as it accumulates figured-out items. Vertical dashed lines mark curriculum transitions where one item is completed and the next begins. The entire curriculum is completed in approximately 32 turns. Scores are representative of observed behavior; exact per-turn values vary across runs.}
  \label{fig:sense_timeline}
\end{figure}

\subsection{POMDP with Internal Reward}
\label{sec:pomdp_v2}

We can frame Sensi v2's learning process as a POMDP augmented with an internal reward signal. Define the agent's epistemic state at time $t$ as:
\begin{equation}
  \mathcal{E}_t = \left(\mathcal{K}_t,\; G_t,\; K_t,\; I^*_t,\; \phi_t,\; \mu_{I^*_t}\right)
  \label{eq:epistemic}
\end{equation}
where $\mathcal{K}_t$ is the fact set, $G_t$ is guesses, $K_t$ is figured-out, $I^*_t$ is the active learning item, $\phi_t$ is the sense score, and $\mu_{I^*_t}$ is the evaluation metric for the active item. The transition dynamics are:
\begin{equation}
  \mathcal{E}_{t+1} = T\!\left(\mathcal{E}_t,\; a_t,\; \mathcal{F}_{t+1},\; \text{LLM}_1(\mathcal{F}_t, \mathcal{F}_{t+1}),\; \text{LLM}_3(\cdot),\; \text{LLM}_4(\cdot)\right)
  \label{eq:transition_dynamics}
\end{equation}
The policy $\pi$ is the composition of all five LLM modules:
\begin{equation}
  a_t = \pi(\mathcal{E}_t, \mathcal{F}_t) = \text{LLM}_5\!\Big(\text{LLM}_4\!\big(\mathcal{F}_t, \Delta_t, \mathcal{E}_t\big)\Big)
  \label{eq:policy}
\end{equation}

What distinguishes this formulation is that the \textbf{reward signal is internal}. The agent does not directly optimize for the game score. Instead, it optimizes for the sense score $\phi_t$ on the current learning item. The game score is an emergent property of learning the correct things in the correct order. This is analogous to intrinsic motivation in reinforcement learning~\citep{schmidhuber1991curious,pathak2017curiosity}, but implemented through LLM-based self-evaluation rather than curiosity-driven exploration bonuses~\citep{sutton1998reinforcement}.

\section{Experiments}
\label{sec:experiments}

\subsection{Setup}

\textbf{Sensi v1} uses ChatGPT~5.1 (Thinking High) as the backbone model for both Observer and Actor roles. \textbf{Sensi v2} uses Gemini~3.1 Pro for all five pipeline stages. Both versions are implemented using DSPy~\citep{khattab2023dspy} for structured LLM interactions, with SQLite for state persistence. The sense score threshold is $\tau = 8/10$ for all learning items.

We note that the use of different backbone models across versions is a confounding factor. This was driven by practical considerations (model availability and multimodal capabilities at time of development) rather than experimental design. We discuss this limitation in \S\ref{sec:limitations}.

\subsection{Baselines}

We compare against the following:
\begin{itemize}[leftmargin=2em]
  \item \textbf{Sensi v1}: The two-player architecture without curriculum learning.
  \item \textbf{Agentica (Symbolica)}: A reported system requiring approximately 1,600--3,000 game interactions per game to build understanding.
  \item \textbf{Random agent}: Uniform random action selection as a lower bound.
\end{itemize}

Our primary comparison axis is \emph{sample efficiency}---the number of game interactions required to complete an exploration or learning phase---rather than win rate alone. This is because sample efficiency determines the practical viability of test-time learning under compute constraints.

\begin{figure}[t]
  \centering
  \begin{tikzpicture}[
    box/.style={rectangle, draw=black!60, thick, minimum width=2.2cm, minimum height=0.9cm, align=center, rounded corners=3pt, font=\small},
    corebox/.style={rectangle, draw=blue!60!black, fill=blue!10, thick, minimum width=2.2cm, minimum height=0.9cm, align=center, rounded corners=3pt, font=\small},
    newbox/.style={rectangle, draw=red!60!black, fill=red!8, thick, minimum width=2.0cm, minimum height=0.8cm, align=center, rounded corners=3pt, font=\scriptsize},
    iobox/.style={rectangle, draw=gray!50, fill=gray!8, minimum width=1.4cm, minimum height=0.6cm, align=center, rounded corners=2pt, font=\scriptsize},
    arr/.style={-{Stealth[length=5pt]}, thick},
  ]

    \node[font=\bfseries] at (1.5,2.2) {Sensi v1};

    \node[iobox] (v1f) at (0,1.2) {$\mathcal{F}_t$};
    \node[corebox] (v1p1) at (1.5,0) {\textbf{Player$_1$}\\[-2pt]{\scriptsize Observer}};
    \node[corebox] (v1p2) at (1.5,-2.0) {\textbf{Player$_2$}\\[-2pt]{\scriptsize Actor}};
    \node[iobox] (v1a) at (3.0,-2.0) {$a_t$};

    \draw[arr] (v1f) |- (v1p1);
    \draw[arr] (v1p1) -- node[right, font=\scriptsize] {$G_t, K_t$} (v1p2);
    \draw[arr] (v1p2) -- (v1a);

    \draw[arr, gray!60, dashed] (v1p2.west) -- ++(-1.2,0) |- (v1p1.west);
    \node[font=\tiny\itshape, text=gray] at (-0.7,-0.7) {next turn};

    \node[font=\bfseries] at (8.0,2.2) {Sensi v2};

    \node[iobox] (v2f) at (5.2,1.2) {$\mathcal{F}_t$};
    \node[newbox] (v2fd) at (6.5,1.2) {FrameDiff\\[-1pt]LLM$_1$};
    \node[newbox] (v2mg) at (8.5,1.2) {MetricGen\\[-1pt]LLM$_2$};
    \node[newbox] (v2ss) at (10.5,1.2) {SenseScore\\[-1pt]LLM$_3$};
    \node[corebox] (v2p1) at (8.0,0) {\textbf{Player$_1$}\\[-2pt]{\scriptsize Observer}};
    \node[corebox] (v2p2) at (8.0,-2.0) {\textbf{Player$_2$}\\[-2pt]{\scriptsize Actor}};
    \node[iobox] (v2a) at (10.5,-2.0) {$a_t$};

    \node[newbox, minimum width=1.8cm, fill=orange!8, draw=orange!60!black] (v2db) at (5.5,-1.0) {\textbf{SQLite}\\[-1pt]DB};

    \draw[arr] (v2f) -- (v2fd);
    \draw[arr] (v2fd) -- (v2mg);
    \draw[arr] (v2mg) -- (v2ss);
    \draw[arr] (v2fd.south) -- ++(0,-0.3) -| (v2p1.north west);
    \draw[arr] (v2ss.south) -- ++(0,-0.2) -| (v2p1.north east);
    \draw[arr] (v2p1) -- node[right, font=\scriptsize] {$G_t, K_t$} (v2p2);
    \draw[arr] (v2p2) -- (v2a);

    \draw[arr, orange!60!black, dashed] (v2db) -- (v2p1);
    \draw[arr, orange!60!black, dashed] (v2db) |- (v2p2.west);
    \draw[arr, orange!60!black, dashed] (v2p1.south west) -- (v2db);

    \draw[thick, gray!30, dashed] (3.8,2.5) -- (3.8,-3.0);

    \node[corebox, minimum width=1.0cm, minimum height=0.4cm, font=\tiny] at (2.0,-3.5) {shared};
    \node[font=\tiny] at (3.2,-3.5) {= v1 core};
    \node[newbox, minimum width=1.0cm, minimum height=0.4cm, font=\tiny] at (5.5,-3.5) {new};
    \node[font=\tiny] at (6.5,-3.5) {= v2 additions};

  \end{tikzpicture}
  \caption{\textbf{Architectural comparison of Sensi v1 and v2.} V1 (left) uses a simple two-player loop with shared hypothesis lists. V2 (right) preserves the Player$_1$/Player$_2$ core (blue) but adds frame differencing, dynamic metric generation, sense scoring, and the database-as-control-plane (red/orange). The two-player core is embedded within a richer learning infrastructure.}
  \label{fig:v1_v2}
\end{figure}
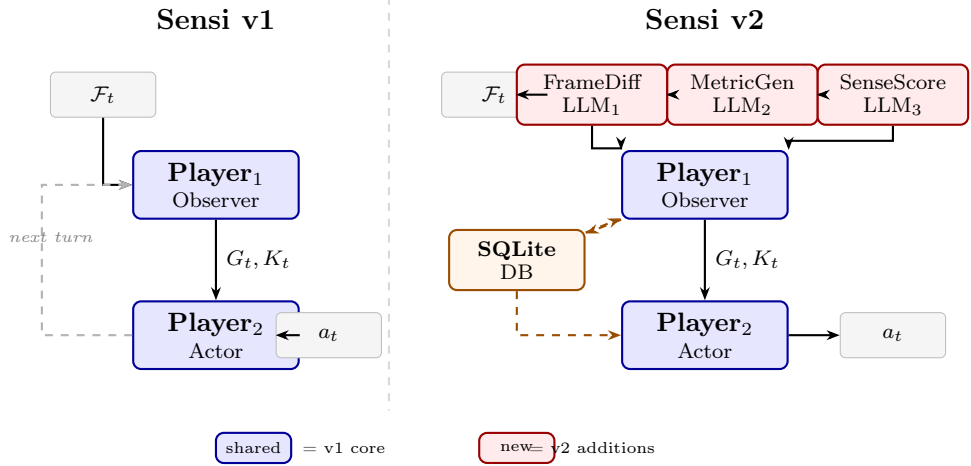

\section{Results and Analysis}
\label{sec:results}

\subsection{Task Performance}

We report task performance honestly and upfront: \textbf{Sensi v1 solves 2 levels; Sensi v2 solves 0 levels.} Table~\ref{tab:comparison} summarizes the comparison.

\begin{table}[t]
  \centering
  \caption{\textbf{Comparison of Sensi versions and baselines.} Sample efficiency is measured as the number of game interactions to complete a learning or exploration phase. V2 achieves dramatically higher sample efficiency despite solving fewer levels.}
  \label{tab:comparison}
  \small
  \begin{tabular}{lccccc}
    \toprule
    \textbf{System} & \textbf{Backbone} & \textbf{Lvls Won} & \textbf{Tries} & \textbf{Curric.} & \textbf{Judge} \\
    \midrule
    Random agent & --- & 0 & --- & \texttimes & \texttimes \\
    Agentica (Symbolica) & Unknown & Unknown & 1,600--3,000 & \texttimes & \texttimes \\
    Sensi v1 & ChatGPT 5.1 & 2 & Variable & \texttimes & \texttimes \\
    Sensi v2 & Gemini 3.1 Pro & 0 & $\sim$32 & \checkmark & \checkmark \\
    \bottomrule
  \end{tabular}
\end{table}

The v2 result---0 levels won---deserves contextualization. The agent \emph{did} complete its entire learning curriculum. It systematically explored what each action does, learned about the energy system, and formed a hypothesis about the win condition. The learning pipeline executed correctly: the state machine transitioned as designed, sense scores rose to meet thresholds, and figured-out items were promoted to facts. The agent learned the wrong things, but it learned them efficiently and systematically.

\subsection{Curriculum Execution}

The curriculum state machine operated correctly across all runs. The agent:
\begin{enumerate}[leftmargin=2em]
  \item Started with the first learning item (``learn what each action does''),
  \item Accumulated guesses and figured-out items about action effects,
  \item Received rising sense scores as the figured-out list grew,
  \item Transitioned to \texttt{completed} when $\phi_t \geq \tau = 8$,
  \item Promoted figured-out items to facts,
  \item Advanced to the next learning item (``learn about energy''),
  \item Repeated the cycle through all curriculum items.
\end{enumerate}

This validates the programmable context architecture: the database-mediated curriculum successfully guided the LLM agent through a structured learning process spanning multiple game resets. The agent's behavior changed appropriately at each curriculum transition, demonstrating that external symbolic control can effectively steer neural inference.

\subsection{Sample Efficiency}
\label{sec:efficiency}

The agent completed its entire learning curriculum in approximately \textbf{32 action attempts}. For comparison, Agentica from Symbolica reportedly requires 1,600 to 3,000 interactions per game:
\begin{equation}
  \text{Sample efficiency ratio} = \frac{N_{\text{baseline}}}{N_{\text{Sensi}}} = \frac{1600\text{--}3000}{32} \approx 50\text{--}94\times
  \label{eq:efficiency}
\end{equation}

This 50--94$\times$ improvement in sample efficiency is the paper's central quantitative result. Even though v2 learned incorrect facts, the \emph{speed} of learning is what matters for the broader research agenda. If an architecture can learn from 32 interactions, the bottleneck is not sample efficiency but \emph{grounding}---ensuring that what the LLM perceives is actually correct. This is a fundamentally different (and more tractable) problem.

In the context of test-time compute research~\citep{yuksekgonul2025tttdiscover}, this is especially relevant. Test-time compute is expensive. An architecture that needs 32 interactions to form hypotheses about a new task is fundamentally more practical than one requiring 3,000, particularly when combined with budget-constrained inference pipelines.

\subsection{Failure Analysis: Self-Consistent Hallucination Cascade}
\label{sec:failure}

The failure mode is instructive and constitutes a diagnostic contribution. The frame differencing step ($\text{LLM}_1$) produces natural language descriptions of visual changes between frames. Gemini~3.1 Pro, when examining small pixel-art frames scaled to $10\times$, frequently misidentifies objects, positions, and movements. A player moving left might be described as ``object shifted right.'' A UI element showing remaining energy might be described as ``decorative border pattern.''

These errors propagate through the entire pipeline in what we term a \textbf{self-consistent hallucination cascade}:
\begin{equation}
  \Delta_t^{\text{wrong}} \;\rightarrow\; K_t^{\text{wrong}} \;\rightarrow\; \phi_t^{\text{high}} \;\rightarrow\; \sigma_{I^*} = \texttt{completed} \;\rightarrow\; \mathcal{K}_{t+1}^{\text{wrong}}
  \label{eq:cascade}
\end{equation}

The cascade operates as follows:
\begin{enumerate}[leftmargin=2em]
  \item \textbf{Perception error.} The frame differencing LLM ($\text{LLM}_1$) produces an incorrect diff $\Delta_t^{\text{wrong}}$.
  \item \textbf{Hypothesis contamination.} Player~1 ($\text{LLM}_4$) builds figured-out items based on the wrong diff, producing $K_t^{\text{wrong}}$.
  \item \textbf{Spurious validation.} The sense scorer ($\text{LLM}_3$) evaluates the agent's understanding based on the \emph{internal consistency} of the figured-out items---not their correspondence to ground truth. Since the items are coherent (just wrong), the scorer assigns a high score $\phi_t^{\text{high}}$.
  \item \textbf{Premature completion.} The high score triggers the state machine transition: $\sigma_{I^*} \leftarrow \texttt{completed}$.
  \item \textbf{Error compounding.} The (wrong) figured-out items are promoted to facts $\mathcal{K}_{t+1}^{\text{wrong}}$, which become part of the context for the next learning item, compounding the error.
\end{enumerate}

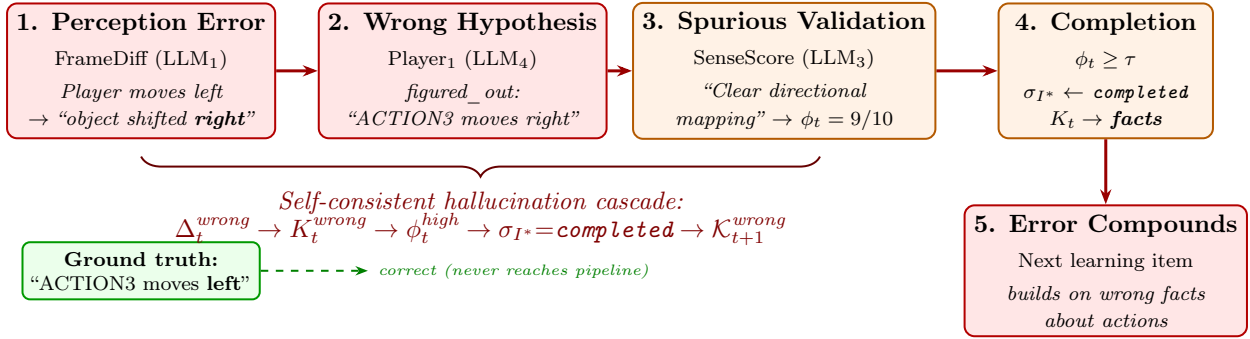
\begin{figure}[t]
  \centering
  \resizebox{\textwidth}{!}{%
  \begin{tikzpicture}[
    step/.style={rectangle, draw=#1!70!black, fill=#1!10, thick, minimum width=3.0cm, minimum height=1.4cm, align=center, rounded corners=4pt, font=\small},
    step/.default=red,
    arr/.style={-{Stealth[length=6pt]}, very thick, color=red!60!black},
    garr/.style={-{Stealth[length=5pt]}, thick, color=green!50!black, dashed},
  ]

    \node[step=red] (s1) at (0,0) {\textbf{1. Perception Error}\\[2pt]{\scriptsize FrameDiff (LLM$_1$)}\\[1pt]{\scriptsize\itshape Player moves left}\\[-1pt]{\scriptsize\itshape $\to$ ``object shifted \textbf{right}''}};

    \node[step=red] (s2) at (4.5,0) {\textbf{2. Wrong Hypothesis}\\[2pt]{\scriptsize Player$_1$ (LLM$_4$)}\\[1pt]{\scriptsize\itshape figured\_out:}\\[-1pt]{\scriptsize\itshape ``ACTION3 moves right''}};

    \node[step=orange] (s3) at (9,0) {\textbf{3. Spurious Validation}\\[2pt]{\scriptsize SenseScore (LLM$_3$)}\\[1pt]{\scriptsize\itshape ``Clear directional}\\[-1pt]{\scriptsize\itshape mapping'' $\to$ $\phi_t = 9/10$}};

    \node[step=orange] (s4) at (13.5,0) {\textbf{4. Completion}\\[2pt]{\scriptsize $\phi_t \geq \tau$}\\[1pt]{\scriptsize\itshape $\sigma_{I^*} \leftarrow$ \texttt{completed}}\\[-1pt]{\scriptsize\itshape $K_t \to$ \textbf{facts}}};

    \node[step=red] (s5) at (13.5,-2.8) {\textbf{5. Error Compounds}\\[2pt]{\scriptsize Next learning item}\\[1pt]{\scriptsize\itshape builds on wrong facts}\\[-1pt]{\scriptsize\itshape about actions}};

    \draw[arr] (s1) -- (s2);
    \draw[arr] (s2) -- (s3);
    \draw[arr] (s3) -- (s4);
    \draw[arr] (s4) -- (s5);

    \node[rectangle, draw=green!60!black, fill=green!8, thick, rounded corners=3pt, minimum width=2.5cm, minimum height=0.8cm, font=\scriptsize, align=center] (truth) at (0,-2.8) {\textbf{Ground truth:}\\``ACTION3 moves \textbf{left}''};
    \draw[garr] (truth.east) -- ++(1.5,0) node[right, font=\tiny\itshape, text=green!50!black] {correct (never reaches pipeline)};

    \draw[decorate, decoration={brace, amplitude=8pt, mirror}, thick, red!40!black] (0,-1.2) -- (9.5,-1.2) node[midway, below=10pt, font=\small\itshape, text=red!50!black, align=center] {Self-consistent hallucination cascade:\\[-1pt]$\Delta_t^{\text{wrong}} \to K_t^{\text{wrong}} \to \phi_t^{\text{high}} \to \sigma_{I^*}{=}\texttt{completed} \to \mathcal{K}_{t+1}^{\text{wrong}}$};

  \end{tikzpicture}%
  }
  \caption{\textbf{Self-consistent hallucination cascade.} A frame differencing error (step~1) propagates through the pipeline. Player$_1$ builds a wrong hypothesis (step~2), which the sense scorer validates because it is \emph{internally consistent} (step~3). This triggers premature completion (step~4), and the wrong knowledge contaminates subsequent learning items (step~5). The ground truth (green, bottom-left) never enters the pipeline.}
  \label{fig:cascade}
\end{figure}

This diagnosis is precise and actionable. The architecture---the curriculum, the state machine, the sense scoring pipeline---all function correctly. The information flow is exactly right. The perception layer is the single point of failure, and it is the most straightforward component to improve through better vision models, hybrid programmatic-neural pixel analysis, or ground-truth anchoring mechanisms.

\section{Discussion}
\label{sec:discussion}

\textbf{Architecture validated independently of task performance.} The 0-level result for v2 might appear to invalidate the approach, but we argue the opposite. The curriculum executed correctly, the state machine transitioned properly, the sense scores rose and fell as expected, and the agent systematically worked through its learning queue. The failure is localized to a single, identifiable component (frame differencing), not to the architectural design. This is analogous to validating a sorting algorithm's correctness even when the comparator function is buggy: the algorithm works; the comparator needs fixing.

\textbf{The bottleneck has shifted.} Before Sensi v2, the question was ``can LLM agents learn game mechanics efficiently?'' After v2, the question becomes ``given that they can learn efficiently, how do we ensure they learn correctly?'' This is a more tractable problem because perceptual grounding admits direct engineering solutions: programmatic pixel comparison, fine-tuned vision models, structured visual analysis, or hybrid approaches that combine LLM perception with deterministic pixel differencing. The curriculum and learning architecture, by contrast, required novel design.

\textbf{Implications for test-time compute budgets.} The 32-try result has practical implications beyond games. Any test-time inference scenario---novel tool use, unfamiliar APIs, new document formats---could benefit from architectures that form structured hypotheses from minimal interactions. A 50--94$\times$ reduction in required interactions translates directly to reduced cost and latency in deployment settings where test-time compute is budgeted.

\textbf{Database-as-control-plane as a general pattern.} We believe the database-as-control-plane pattern is underexplored in the agent architecture literature. Most agent frameworks manage context through prompt templating---static templates with variable substitution. Sensi v2 demonstrates that externalizing cognitive state to a database and programmatically injecting it into prompts creates more modular, steerable, and debuggable agents. This pattern naturally supports multi-agent coordination, human-in-the-loop oversight, and post-hoc analysis of agent behavior through standard database queries.

\section{Related Work}
\label{sec:related}

\textbf{Test-time computation and training.} The spectrum of test-time adaptation ranges from inference-time reasoning~\citep{wei2022chain,wang2023selfconsistency} through in-context learning to full gradient-based test-time training~\citep{sun2024learning,yuksekgonul2025tttdiscover}. TTT-Discover~\citep{yuksekgonul2025tttdiscover} is most relevant to our work: it performs reinforcement learning on a single test problem, training the model's weights to discover solutions. Sensi differs fundamentally in that no gradient updates occur; instead, learning is realized through structured knowledge accumulation in an external database. The framework applies sequential skill acquisition at inference time (no retraining) to improve stability, out-of-distribution generalization, and continual adaptation in agentic workflows. This directly extends ideas in EvoTest~\citep{he2025evotest}, an evolutionary test-time learning framework for self-improving systems. This makes Sensi applicable to black-box LLM APIs where weight access is unavailable.

\textbf{LLM agents for games and environments.} Voyager~\citep{wang2024voyager} builds an open-ended embodied agent with a skill library, curriculum, and code-as-action paradigm in Minecraft. Unlike Sensi, Voyager operates in a well-understood environment with known mechanics. ReAct~\citep{yao2023react} interleaves reasoning and acting but does not maintain persistent state across episodes. Reflexion~\citep{shinn2024reflexion} adds verbal reinforcement learning through self-reflection, which is conceptually similar to Sensi's sense scoring but without the curriculum structure or external state machine. SPRING~\citep{wu2024spring} uses game manuals as grounding documents---the opposite of Sensi's approach, which assumes no prior game knowledge.

\textbf{Curriculum learning.} The idea of ordering training examples by difficulty dates to Bengio et al.~\citep{bengio2009curriculum} and has been extensively studied in deep RL~\citep{portelas2020automatic}. Sensi's contribution is applying curriculum learning at \emph{test time} to structure an LLM agent's exploration, rather than ordering training data during model development. The curriculum is hand-designed in the current work; automatic curriculum generation~\citep{portelas2020automatic} is a natural extension.

\textbf{LLM self-evaluation.} The use of LLMs as judges has been studied extensively~\citep{zheng2024judging}, primarily for evaluating model outputs in benchmarking contexts. Self-Refine~\citep{madaan2024selfrefine} uses self-feedback for iterative improvement. Sensi's sense scorer extends this paradigm in two ways: (1)~the evaluation rubric is itself generated by an LLM, making the system adaptive to arbitrary learning domains, and (2)~the score is used as a control signal for a state machine rather than just feedback for refinement.

\textbf{ARC and abstraction reasoning.} The ARC benchmark~\citep{chollet2019measure} tests abstract reasoning through visual pattern completion. Program synthesis approaches~\citep{johnson2021program} attempt to solve ARC tasks by generating programs. ARC-AGI-3~\citep{arcprize2024arcagi} extends this to interactive game-playing, requiring online learning rather than one-shot inference. Sensi is, to our knowledge, the first architecture to apply curriculum-based test-time learning specifically to the ARC-AGI-3 game-playing challenge.

\section{Limitations and Future Work}
\label{sec:limitations}

\textbf{Limitations.} Several limitations should be noted: (1)~Sensi v2 won 0 game levels, meaning the architecture has not yet demonstrated end-to-end task success with curriculum learning. (2)~The curriculum is hand-designed; the choice and ordering of learning items requires human judgment about what is learnable and in what order. (3)~V1 and v2 use different backbone models (ChatGPT~5.1 and Gemini~3.1 Pro, respectively), confounding direct comparison of architectural changes with model capability differences. (4)~The sample efficiency comparison to Agentica/Symbolica is based on reported numbers rather than controlled experiments with identical conditions. (5)~The sense scorer evaluates internal consistency rather than ground-truth correctness, creating a fundamental vulnerability to coherent hallucinations.

\textbf{Future work.} The most immediate priority is \textbf{fixing the perception layer}. Candidate approaches include: (i)~hybrid programmatic and LLM-based frame differencing, where deterministic pixel comparison provides ground truth that the LLM interprets; (ii)~ground-truth anchoring for the sense scorer, where a subset of evaluation criteria are verified against actual game state rather than relying solely on LLM judgment; and (iii)~fine-tuned vision models specifically trained on pixel-art game frames.

Beyond perception, several directions are promising: \textbf{automatic curriculum generation}, where the agent determines its own learning order based on initial exploration; \textbf{cross-game transfer}, where facts learned in one game inform hypotheses in structurally similar games; and \textbf{multi-agent curriculum learning}, where multiple agents with different curricula share and validate knowledge through the database-as-control-plane.

\section{Conclusion}
\label{sec:conclusion}

We presented Sensi, an LLM agent architecture that evolves across two iterations to address the sample efficiency problem in test-time game learning. Sensi v1 demonstrates that separating perception from action via a two-player architecture enables systematic hypothesis accumulation, solving 2 ARC-AGI-3 levels. Sensi v2 adds curriculum learning, a database-as-control-plane, and LLM-as-judge evaluation, achieving 50--94$\times$ greater sample efficiency than comparable systems---completing its learning curriculum in approximately 32 interactions versus 1,600--3,000 for baselines.

V2's 0-level result is an honest negative finding, but the failure mode is precisely diagnosed: a self-consistent hallucination cascade originating in the perception layer, where frame differencing errors produce internally coherent but factually incorrect knowledge that passes the sense scorer's consistency checks. This diagnosis shifts the research bottleneck from ``how to learn efficiently'' to ``how to perceive correctly''---a more tractable engineering problem.

The broader contribution is methodological: structured test-time learning---with curriculum, state machine, and programmable context---works as an architectural pattern. The 32-try budget suggests that LLM agents can form structured hypotheses about unknown environments with human-like sample efficiency, provided the perception layer is reliable. Fixing perception is the next step; the learning architecture is ready.

\section*{Acknowledgments}

The author thanks the ARC Prize Foundation for creating the ARC-AGI-3 challenge. The Sensi agent code is built on the ARC-AGI-3-Agents framework.\footnote{\url{https://github.com/arcprize/ARC-AGI-3-Agents}} This work was conducted independently without institutional funding.

\bibliographystyle{plainnat}
\bibliography{references}

\appendix

\section{Sensi v1: Discovered Game Knowledge}
\label{app:v1_knowledge}

For completeness, we list the 15 figured-out items that Sensi v1 discovered for game LS20, Level~1. These items were accumulated over the course of gameplay and represent the agent's complete model of the game mechanics:

\begin{enumerate}[leftmargin=2em]
  \item RESET starts the game.
  \item Blue platform (with red top) is our player.
  \item ACTION1 moves the player 1 pixel up.
  \item ACTION2 moves the player 1 pixel down.
  \item ACTION3 moves the player 1 pixel left.
  \item ACTION4 moves the player 1 pixel right.
  \item Each action consumes 1 unit of energy shown as dots.
  \item Key generator produces matching keys for doors.
  \item Walking into a key generator creates a key of that color.
  \item Keys are consumed when used to open matching doors.
  \item Doors disappear when opened with the correct key.
  \item Energy dots provide additional energy when collected.
  \item Game over occurs when energy reaches zero.
  \item Stars are collectible items.
  \item Collecting all stars completes the level.
\end{enumerate}

These items were discovered purely through interaction, without any prior game knowledge. The accuracy of these items (verified against actual game mechanics) demonstrates that the two-player architecture can produce correct game models when the perception layer functions adequately.

\section{Algorithm: Sensi v2 Turn Execution}
\label{app:algorithm}

\begin{algorithm}[H]
\caption{Sensi v2: Single Turn Execution}
\label{alg:turn}
\begin{algorithmic}[1]
\Require Game frame $\mathcal{F}_t$, previous frame $\mathcal{F}_{t-1}$, database DB
\Ensure Action $a_t$

\State $\Delta_t \gets \text{FrameDiff}(\mathcal{F}_{t-1}, \mathcal{F}_t)$ \Comment{LLM$_1$: visual differencing}
\State $I^* \gets \text{DB.get\_active\_item()}$ \Comment{Current learning item}
\State $\mathcal{K}_t \gets \text{DB.get\_facts()}$ \Comment{Accumulated facts}

\If{$\sigma_{I^*} = \texttt{not\_reached}$}
    \State $\sigma_{I^*} \gets \texttt{learning}$
    \State $\mu_{I^*} \gets \text{MetricGen}(I^*)$ \Comment{LLM$_2$: generate rubric}
    \State $\text{DB.store\_metric}(I^*, \mu_{I^*})$
\EndIf

\State $\phi_t, r_t \gets \text{SenseScore}(I^*, \mu_{I^*}, \mathcal{K}_t, K_{t-1})$ \Comment{LLM$_3$: evaluate}

\If{$\phi_t \geq \tau_{I^*}$}
    \State $\sigma_{I^*} \gets \texttt{completed}$
    \State $\text{DB.promote\_to\_facts}(K_{t-1})$ \Comment{Knowledge accumulation}
\EndIf

\State $G_t, K_t \gets \text{Player}_1(\mathcal{F}_t, \Delta_t, \mathcal{K}_t, G_{t-1}, K_{t-1}, I^*, r_t)$ \Comment{LLM$_4$}
\State $\text{DB.store}(G_t, K_t)$

\State $d_t, a_t \gets \text{Player}_2(G_t, K_t, \mathcal{K}_t, I^*)$ \Comment{LLM$_5$}
\State $\text{DB.log\_action}(a_t, d_t)$

\State \Return $a_t$
\end{algorithmic}
\end{algorithm}

\end{document}